\documentclass[runningheads]{llncs}

\usepackage[T1]{fontenc}
\usepackage{graphicx}
\usepackage{booktabs}
\usepackage{array}
\usepackage{multirow}
\usepackage{enumitem}
\usepackage{amssymb}
\usepackage{xcolor}
\usepackage{tikz}
\usetikzlibrary{arrows.meta,backgrounds,fit,positioning}
\usepackage{hyperref}
\hypersetup{
  colorlinks=true,
  linkcolor=blue!60!black,
  citecolor=blue!60!black,
  urlcolor=blue!60!black
}
\begin{document}
\raggedbottom

\title{Decision-Aware Memory Cards: Counterfactual-Inspired Context Selection and Compression for Tool-Using LLM Agents}
\titlerunning{Decision-Aware Memory Cards for LLM Agents}

\author{Xinyu Guan\inst{1} \and Qianyang Zhao\inst{1} \and Yuming Deng\inst{1}}
\authorrunning{X. Guan et al.}
\institute{Alibaba Group, China\\
\email{guanhan.gxy@alibaba-inc.com}\\
\email{zhaoqianyang.zqy@alibaba-inc.com}\\
\email{finaldreamer@qq.com}\\
Corresponding author: Xinyu Guan.}

\maketitle

\begin{abstract}
Modern large language model (LLM) agents do not simply need longer contexts;
they need decision-relevant evidence at the moment of action. We study decision-aware
context selection: ranking retrieved files, tests, traces, rules, and memories
by their expected effect on an agent's next action rather than by semantic
similarity alone. We present the Counterfactual-Inspired Context Layer (CICL),
which builds an instance context graph, estimates decision-oriented utility for
candidate units, and compresses selected evidence into typed memory cards. The
same schema can be instantiated with hosted LLM judges, local surrogates, or
lightweight rankers, making the selection protocol auditable across model
choices. On 50 SWE-bench Verified file-retrieval instances, Qwen3.6-Plus
reranking of BM25 top-50 candidates improves hit@1 from $0.58$ to $0.78$ and
MRR@10 from $0.634$ to $0.790$, with all 2,500 judgments parseable. Controlled
diagnostics show that CICL identifies action-critical evidence: removing the
top-utility semantic unit reduces F1 from $0.245$ to $0.000$. In
selected-then-compressed mode, memory cards save $44.93$ tokens per query while
preserving selected evidence. CICL provides a practical layer for measuring,
ranking, and compressing decision-critical context for tool-using agents. Code
is available at \url{https://github.com/stephen-guan-researcher/CICL}.

\keywords{LLM agents \and Context selection \and Counterfactual utility \and
Memory compression \and LLM-as-a-judge diagnostics.}
\end{abstract}

\section{Introduction}
\label{sec:intro}

A coding agent can fail even when the answer is already in the repository. The
missing clue may be a failing test, a file-local convention, a stale trace to
ignore, or a short rule buried among irrelevant snippets. More context does not
fix this by itself. What matters is the evidence the agent uses before acting:
the piece of context that changes its next command, edit, or test.

\begin{figure}[!t]
\centering
\small
\begingroup
\definecolor{ciclInk}{HTML}{202939}
\definecolor{ciclLine}{HTML}{566273}
\definecolor{ciclMuted}{HTML}{66717C}
\definecolor{ciclBorder}{HTML}{D8DEE4}
\definecolor{ciclGold}{HTML}{FFF2CC}
\definecolor{ciclRose}{HTML}{FBE5E4}
\definecolor{ciclViolet}{HTML}{EEE8FA}
\definecolor{ciclGreen}{HTML}{E7F3E7}
\definecolor{ciclCardLine}{HTML}{C78D86}
\definecolor{ciclCardRule}{HTML}{DFC0BC}
\definecolor{ciclGraphLine}{HTML}{8BA29D}
\definecolor{ciclGraphNode}{HTML}{59877D}
\definecolor{ciclGraphCore}{HTML}{34766C}
\definecolor{ciclUtilityLine}{HTML}{DDD0AA}
\definecolor{ciclAgentLine}{HTML}{9984B6}
\definecolor{ciclLedger}{HTML}{263B4B}
\definecolor{ciclLedgerText}{HTML}{D9E8ED}
\begin{tikzpicture}[
  x=1bp,y=-0.8bp,
  font=\fontencoding{T1}\fontfamily{phv}\selectfont
    \fontsize{7.5bp}{9.8bp}\selectfont,
  text=ciclInk,
  every node/.style={inner sep=0pt,outer sep=0pt,align=center},
  panel/.style={draw=ciclBorder,fill=white,rounded corners=3bp,
    line width=0.6bp},
  title/.style={font=\fontencoding{T1}\fontfamily{phv}\selectfont
    \bfseries\fontsize{8bp}{10.3bp}\selectfont},
  field/.style={anchor=west,font=\fontencoding{T1}\fontfamily{phv}\selectfont
    \bfseries\fontsize{8.5bp}{10.8bp}\selectfont},
  detail/.style={anchor=west,text=ciclMuted,
    font=\fontencoding{T1}\fontfamily{phv}\selectfont
    \fontsize{8bp}{10.3bp}\selectfont},
  utility/.style={font=\fontencoding{T1}\fontfamily{phv}\selectfont
    \fontsize{8.5bp}{10.8bp}\selectfont},
  flow/.style={-{Latex[length=4bp,width=3.6bp]},draw=ciclLine,line width=0.65bp},
  audit/.style={flow,dash pattern=on 2.2bp off 2.2bp}
]
\path[use as bounding box] (0,0) rectangle (346,436);

\draw[panel,fill=ciclGreen] (0,0) rectangle (346,48);
\node[title] at (173,10) {Judge router: one judge per run};
\foreach \x/\name in {3/{Claude Opus 4.7},88.75/{Qwen3.6 Plus},
  174.5/{GPT-5.5},260.25/{Qwen3.5-9B QLoRA}} {
  \draw[panel,rounded corners=2bp] (\x,23) rectangle ++(82.75,18);
  \node at (\x+41.375,32) {\name};
}

\node[title] at (50,82) {Instance evidence};
\draw[draw=ciclLine,fill=white,line width=0.6bp] (16,136) circle[radius=9.5bp];
\node at (16,136) {task};
\draw[draw=ciclLine,line width=0.5bp] (25.5,136) -- (30,136);
\foreach \y/\name in {104/files,120/tests,136/traces,152/rules,168/memory} {
  \draw[draw=ciclLine,line width=0.55bp]
    (30,136) .. controls (36,136) and (34,\y) .. (41,\y);
  \draw[panel,rounded corners=2bp,line width=0.5bp]
    (41,\y-6) rectangle ++(51,12);
  \node at (66.5,\y) {\name};
}
\fill[ciclLine] (30,136) circle[radius=1.2bp];
\draw[flow] (50,174) -- (50,188);

\node[title] at (50,198) {Context graph};
\foreach \id/\x/\y in {0/26/231.51,1/39.8/219.63,2/58.4/218.64,
  3/73.4/229.2,4/65/245.04,5/42.8/245.04,6/49.4/231.51}
  \coordinate (g\id) at (\x,\y);
\foreach \a/\b in {0/1,1/2,2/3,3/4,4/5,5/0,0/6,1/6,2/6,3/6,4/6,5/6}
  \draw[draw=ciclGraphLine,line width=0.6bp] (g\a) -- (g\b);
\foreach \id in {0,1,2,3,4,5}
  \draw[draw=ciclGraphNode,fill=white,line width=0.6bp]
    (g\id) circle[radius=2.4bp];
\draw[draw=ciclGraphNode,fill=ciclGraphCore,line width=0.6bp]
  (g6) circle[radius=3.2bp];
\draw[flow] (50,251) -- (50,267);

\draw[panel,fill=ciclGold] (0,267) rectangle (100,367);
\node[title,font=\fontencoding{T1}\fontfamily{phv}\selectfont
  \bfseries\fontsize{9bp}{11.3bp}\selectfont] at (50,280) {Decision utility};
\foreach \x/\y/\label in {7/299/{action\\shift},52/299/{outcome\\uplift},
  7/331/necessity,52/331/{risk / cost}} {
  \draw[panel,draw=ciclUtilityLine,rounded corners=2bp,line width=0.5bp]
    (\x,\y) rectangle ++(41,28);
  \node[utility] at (\x+20.5,\y+14) {\label};
}

\draw[panel,rounded corners=4bp,draw=ciclCardLine] (127,86) rectangle (256,339);
\draw[panel,rounded corners=4bp,fill=ciclRose,draw=ciclCardLine]
  (123,90) rectangle (252,343);
\node[title,font=\fontencoding{T1}\fontfamily{phv}\selectfont
  \bfseries\fontsize{10bp}{12.3bp}\selectfont] at (187.5,108) {Memory card};
\node[text=ciclMuted] at (187.5,124) {typed schema};
\foreach \y/\name/\ciclFieldDetail in {
  149/Trigger/{When it applies},
  187/Evidence/{Supporting context},
  225/{Action hint}/{Suggested next step},
  263/{Failure if ignored}/{Risk to avoid},
  301/Scope/{Validity boundary}} {
  \node[field] at (134,\y) {\name};
  \node[detail] at (134,\y+13) {\ciclFieldDetail};
}
\foreach \y in {174,212,250,288}
  \draw[draw=ciclCardRule,line width=0.5bp] (134,\y) -- (241,\y);

\draw[panel,fill=ciclViolet] (276,187) rectangle (346,263);
\node[title,font=\fontencoding{T1}\fontfamily{phv}\selectfont
  \bfseries\fontsize{8.5bp}{10.8bp}\selectfont] at (311,203) {Budgeted\\agent};
\draw[panel,draw=ciclAgentLine] (294,220) rectangle (328,245);
\draw[draw=ciclAgentLine,line width=0.5bp] (294,227) -- (328,227);
\foreach \x in {299,303,307}
  \fill[ciclAgentLine] (\x,223.7) circle[radius=0.7bp];
\draw[draw=ciclAgentLine,line width=0.85bp] (301,232) -- (304,235) -- (301,238);
\draw[draw=ciclAgentLine,line width=0.85bp] (308,238) -- (316,238);

\draw[flow] (100,326) -- (123,326);
\draw[flow] (256,225) -- (276,225);
\draw[flow] (113,48) -- (113,284) -- (100,284);

\draw[panel,draw=ciclLedger,fill=ciclLedger,rounded corners=4bp,line width=0.7bp]
  (87,395) rectangle (299,433);
\node[title,text=white,font=\fontencoding{T1}\fontfamily{phv}\selectfont
  \bfseries\fontsize{9bp}{11.3bp}\selectfont] at (193,407) {Evidence ledger};
\node[text=ciclLedgerText,font=\fontencoding{T1}\fontfamily{phv}\selectfont
  \fontsize{8bp}{10.3bp}\selectfont] at (193,422) {supported / limited / deferred};
\draw[audit] (50,367) -- (50,414) -- (87,414);
\draw[audit] (311,263) -- (311,414) -- (299,414);

\end{tikzpicture}
\endgroup
\caption{CICL pipeline. The framework turns instance evidence into a graph,
routes judge signals through a decision-utility engine, and packs memory cards
for a budgeted agent.}
\label{fig:method-pipeline}
\end{figure}
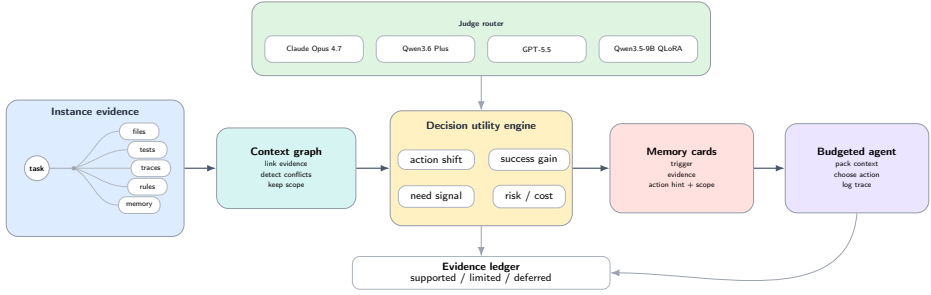

This issue becomes sharper as LLM agents move from text generation to tool use.
Modern agents can interleave reasoning and action \cite{react2023}, call
external~tools~\cite{toolformer2023}, revise behaviour from feedback
\cite{reflexion2023}, and accumulate reusable skills \cite{voyager2023}. In
software engineering, systems such as SWE-agent \cite{sweagent2024} expose
models to repositories, shell commands, tests, and edits. These abilities make
context selection more important, not less: an agent acting on the wrong
evidence can fail quickly and confidently. Recent coding-agent context
benchmarks make this bottleneck explicit
\cite{contextbench2026,swecontextbench2026}.

We therefore study context selection as a decision problem. Standard retrieval
asks which text is most similar to the task. An agent instead needs evidence
that can affect the next step: by changing an action, supporting a better
outcome, becoming necessary for success, or avoiding a harmful choice. CICL
instantiates this idea as a budgeted context layer that separates three roles:
candidate generation, decision-oriented scoring, and compact context packaging.
The same schema can be instantiated with hosted LLMs, local surrogates, or lightweight
rankers, so the protocol remains auditable across model choices.
Figure~\ref{fig:method-pipeline} gives an overview.

The paper asks whether this decision-aware signal helps rank agent context and
whether the schema remains usable when the judge changes. We test this with
SWE-bench Verified file retrieval, controlled removal diagnostics, compression
experiments, and judge-agreement checks. The results show gains in real-code
file retrieval and token-efficient context packing, while also exposing clear
limits: strong baselines remain competitive, and the evidence is file-level
rather than end-to-end patch success. This positioning keeps the claims safely
focused on context selection rather than full agent repair or deployment.

\paragraph{Contributions.}
(1) We frame agent context selection as a decision-time intervention and
formalise a four-component utility for action-critical evidence.
(2) We introduce decision-aware memory cards and a graph-based assembly
pipeline for packing action-oriented context.
(3) We evaluate the same judgment schema across hosted LLMs, Qwen-family
judges, lightweight rankers, and a local Qwen-QLoRA surrogate.
(4) We provide controlled diagnostics identifying where the framework works,
where LLM-based reranking improves open benchmark retrieval, and where
baselines remain stronger.

\section{Related Work}
\label{sec:related}

LLM agent research has broadened evaluation from isolated responses to systems
that coordinate actions, use tools, and preserve state over time. AutoGen and ChatDev
show how model roles can be organised
into collaborative software-development workflows
\cite{autogen2023,chatdev2023}. For repository-level coding, Agentless and
OpenHands cover complementary designs: strong non-agent repair pipelines and
open infrastructure for executing coding agents
\cite{agentless2024,openhands2024}. Memory-oriented benchmarks show
that agent state must be tested across sessions, turns, and self-evolving
updates \cite{memoryarena2026,memoryagentbench2026,evomembench2026}. These
works establish settings where available context matters, but they largely
evaluate complete agent systems or benchmark outcomes. CICL treats this
pre-action context choice as the unit of study rather than evaluating a
complete agent architecture.

External context for agents is usually obtained through retrieval and then
placed in the prompt. Dense retrieval and vector-search systems provide
scalable candidate generation \cite{contriever2022,faiss2019}. RAG conditions
generation on retrieved evidence \cite{rag2020}, while Atlas and HyDE improve
retrieval through retrieval-scale pretraining and hypothetical documents
\cite{atlas2023,hyde2023}. Self-RAG makes this process more selective by
reflecting on when retrieved passages are useful \cite{selfrag2024}. Long-context
evaluations add a complementary caution: information included in a long input
may still be missed or underused \cite{lostinthemiddle2024,longbench2023}.
These methods improve access to evidence or test whether evidence can be used
once supplied. CICL operates one step earlier in the pipeline, at the ranking
stage that decides which retrieved units should reach the agent.

Context management for long-horizon agents creates a related selection pressure.
AutoContext, ACE, and ACON adapt, evolve, or compress context for agentic
settings \cite{autocontext2026,ace2026,acon2025}. Prompt compression methods
reduce input length by preserving salient tokens, spans, or segments
\cite{llmlingua2023,longllmlingua2024,selectivecontext2023}. Contribution-aware
methods ask which units help or harm downstream behaviour: CMI studies causal
memory selection \cite{cmi2026}, and RepoShapley estimates repository-level
contribution for code completion \cite{reposhapley2026}. CICL uses these
mechanisms for a narrower purpose: selecting compact context units whose value
is measured by their role in the agent's decision process, not only by
relevance or compression salience.

\section{Method}
\label{sec:method}

\subsection{Decision-Aware Context Selection}

Consider an agent policy $\pi$ acting on tasks $x \in \mathcal{X}$. At each
decision step the agent receives a context block $C \subseteq \mathcal{U}$,
where $\mathcal{U}$ is the pool of candidate units produced by the instance
graph. Unlike BM25 \cite{bm25}, DPR \cite{dpr2020}, or ColBERT
\cite{colbert2020}, CICL treats selected evidence as an intervention.
A relevance selector solves
$C^{\mathrm{rel}} = \arg\max_{\mathrm{tok}(C) \leq B}
\sum_{c \in C} \mathrm{sim}(c, x)$ under token budget $B$. CICL replaces
$\mathrm{sim}$ with a decision-time utility $U(c,x)$ that estimates whether
$c$ would change the next action or expected success under the agent policy.

\subsection{Counterfactual-Inspired Utility}

Let $C^-$ denote the context assembled before considering candidate $c$, and
let $C^+ = C^- \cup \{c\}$. With $\pi(a \mid x, C)$ denoting the conceptual
next-action distribution under context $C$, CICL decomposes utility into four
components. For hosted LLM agents, we do not assume direct access to this
distribution: $\pi$ defines the target quantities, while CICL estimates the
components using simulator probes, structured LLM judgments, or learned rankers
under the same schema:
\begin{equation}
\begin{array}{rcl}
\Delta_{\mathrm{act}}(c, x) &=& \mathbb{E}\big[\,\mathbf{1}\{\arg\max_a \pi(a \mid
x, C^+) \neq \arg\max_a \pi(a \mid x, C^-)\}\,\big] \\
\Delta_{\mathrm{out}}(c, x) &=& \mathbb{E}\big[\,V(x, C^+) - V(x, C^-)\,\big] \\
N(c, x) &=& \Pr\big[\,\mathrm{success}(x,C^+) = 1 \wedge
\mathrm{success}(x,C^-) = 0\,\big] \\
R(c, x) &=& \Pr\big[\,c \mbox{ induces negative transfer on } x \,\big].
\end{array}
\end{equation}
Here, $V$ denotes an expected success score. $N(c,x)$ is a necessity-style
proxy for cases where adding $c$ changes failure into success, and $R(c,x)$
captures negative transfer. CICL aggregates the components with fixed signed
weights:
\begin{equation}
U(c, x) = \alpha\,\Delta_{\mathrm{act}}(c, x) + \beta\,\Delta_{\mathrm{out}}(c, x)
+ \gamma\,N(c, x) - \lambda\,R(c, x).
\label{eq:utility}
\end{equation}

For judge-style outputs, let hats denote evaluator-produced estimates of the
corresponding conceptual components. Eq.~\ref{eq:judgment-to-score} is a fixed
operational instantiation of Eq.~\ref{eq:utility}: it plugs the judge fields
into the same signed utility structure and adds a bounded cost penalty:
\begin{equation}
\begin{array}{rcl}
s(c,x) &=& 0.34\,\widehat{\Delta}_{\mathrm{act}}(c,x)
+ 0.26\,\widehat{N}(c,x)
+ 0.28\,\widehat{\Delta}_{\mathrm{out}}(c,x) \\
&& {} - 0.22\,\widehat{R}(c,x)
- 0.08\,\widehat{\mathrm{cost}}(c,x).
\end{array}
\label{eq:judgment-to-score}
\end{equation}
In the implementation,
$\widehat{\mathrm{cost}}(c,x)=
\min(1,\mathrm{tok}(c)/1000+0.2\widehat{R}(c,x))$ for
LLM-judged units, matching the released scorer. We keep this aggregation fixed
across Claude Opus, Qwen, and GPT-5.5 via Codex judge runs to prevent ablation
drift; deterministic proxy ablations use the same component signs with
model-free estimates. The coefficients in Eq.~\ref{eq:judgment-to-score} are
pre-set heuristic weights, not learned from test labels. Component-removal
ablations provide the current sensitivity evidence; equal-weight,
random-weight, and learned-weight sweeps remain future work.
Table~\ref{tab:judgment-schema} lists the judge output fields. We stress that
``causal'' here denotes counterfactual-inspired utility estimation rather than
formal causal identification: the expectations above are not identified by any
randomised intervention, and we defer a detailed discussion of this boundary to
Section~\ref{sec:limitations}.

\begin{table}[!htbp]
\centering
\caption{Eight-field judge schema used by Claude Opus and Qwen. The four
utility fields estimate the hatted terms in Eq.~\ref{eq:judgment-to-score};
confidence is retained for audit and diagnostics, and token cost comes from
context metadata.}
\label{tab:judgment-schema}
\scriptsize
\setlength{\tabcolsep}{4pt}
\renewcommand{\arraystretch}{0.92}
\begin{tabular}{@{}llll@{}}
\toprule
Field & Role & Field & Role \\
\midrule
\texttt{no\_context\_action} & action without unit &
\texttt{with\_context\_action} & action with unit \\
\texttt{action\_shift} & action change $[0,1]$ &
\texttt{necessity} & dependence $[0,1]$ \\
\texttt{expected\_outcome\_uplift} & value gain $[0,1]$ &
\texttt{negative\_transfer\_risk} & conflict risk $[0,1]$ \\
\texttt{reason} & short rationale &
\texttt{confidence} & self-confidence $[0,1]$ \\
\bottomrule
\end{tabular}
\end{table}

\subsection{Instance Context Graph}

For each repository or environment instance, CICL constructs a graph whose
nodes correspond to files, symbols, task memories, rules, failures, and
strategy records. Edges capture containment, similarity, conflict,
precondition, and task-memory relations. The graph combines lexical and
structural retrieval with one-hop neighbour expansion, which supports recovery
of decision-relevant context even when lexical overlap with the query is weak.
Each node is a context unit annotated with an identifier,
instance id, type, source, content, token cost, and confidence score.
Importantly, the graph never requires gold context identifiers for
selection; gold ids appear only during offline evaluation and in oracle
baselines. We audit all method-facing artifacts for gold-label leakage and
include the audit script in the reproducibility package.

\subsection{Decision-Aware Memory Cards}

CICL compiles selected units into compact memory cards with five mandatory
fields---\emph{trigger} (when to consult), \emph{evidence} (supporting
clue), \emph{action hint} (next-action verb), \emph{failure-if-ignored}
(risk if skipped), and \emph{scope} (applicable boundary)---plus a
diagnostic \emph{decision-utility estimate} for ordering. The format
prioritises decision usefulness over exhaustive semantic fidelity. Generic
prompt compression often optimises token- or sentence-level importance,
as in LLMLingua \cite{llmlingua2023} and LongLLMLingua
\cite{longllmlingua2024}, or filters by salience \cite{selectivecontext2023};
CICL stores typed decision
fields. A deterministic structural audit checks required-field completeness,
action-verb presence, compression ratio, and absence of placeholder text.

\subsection{Budget-Aware Assembly}

At inference time, CICL retrieves candidate units, expands graph neighbours,
scores candidates via $U$ or its operational estimate $s$, and packs the
highest-utility evidence under a fixed token budget. We distinguish post-selection compression, which first
selects ids and then compresses their text, from pre-budget compression,
which changes candidate costs before packing and can therefore change the
selected ids. We report the two modes separately to avoid conflating
compression gains with changes in selection.

\subsection{Opus-Assisted Annotation and an Open-Weights Surrogate Judge}

Claude Opus 4.7 is used to assist a human-designed context-utility annotation
workflow. These provider-assisted diagnostic annotations are not human gold
labels; they train small CICL context rankers rather than a reproduced teacher:
a 25-feature pairwise linear ranker and a two-layer MLP. All features are
available at inference time, including proxy utility components, retrieval and
lexical-overlap scores, unit type, token cost, confidence, history success,
and graph signals such as conflict degree and
source/task overlap. Rankers are trained from within-task positive--negative
candidate pairs, following preference-learning style supervision used in
instruction tuning \cite{instructgpt2022} and summarisation from preferences
\cite{stiennon2020}, and are then used only for context ordering. After
excluding features unavailable at inference time, the cleaned linear ranker
reaches $0.939$ pairwise accuracy on the v1 suite. Provider-side
\texttt{llm\_*} features are not used by any reported selector.

As a local open-weights alternative we fine-tune Qwen3.5-9B with QLoRA on two
16\,GB V100 GPUs for the same eight-field schema, following LoRA adaptation
\cite{lora2022} and QLoRA quantized fine-tuning \cite{qlora2023}. We release
the adapter at
{\footnotesize\url{https://huggingface.co/XinyuGuan/CICL}}.
It must be loaded with
\texttt{Qwen/Qwen3.5-9B} and is used here as an agreement surrogate rather than
a standalone judge or a Claude Opus replacement. Training uses the released v1
Opus-assisted SFT split ($1{,}400$ examples; $1{,}256$ train and $144$
validation examples, split by task); evaluation measures selection-level
agreement via top-$k$ Jaccard and Spearman $\rho$ on 710 candidates from 25 base
tasks. This keeps deterministic, Opus-assisted, lightweight-ranker, and
Qwen-surrogate scores auditable as separate components.

\section{Experimental Setup}
\label{sec:setup}

\subsection{Tasks and Benchmarks}

Table~\ref{tab:dataset-summary} gives the minimal data map. SWE-bench Verified
is the main real-code retrieval benchmark \cite{swebench2024}; synthetic suites
isolate mechanism; RepoBench-R tests compression \cite{repobench2023}.
CodeSearchNet motivates the broader semantic code-search setting that these
repository-level retrieval tasks inherit \cite{codesearchnet2019}.

\begin{table}[!htbp]
\centering
\caption{Dataset summary with scale, use, and evaluation scope.}
\label{tab:dataset-summary}
\scriptsize
\setlength{\tabcolsep}{3.5pt}
\renewcommand{\arraystretch}{1.0}
\begin{tabular}{@{}>{\raggedright\arraybackslash}p{0.22\linewidth}
                >{\raggedright\arraybackslash}p{0.20\linewidth}
                >{\raggedright\arraybackslash}p{0.21\linewidth}
                >{\raggedright\arraybackslash}p{0.27\linewidth}@{}}
\toprule
Source & Use & Scale & Evaluation scope \\
\midrule
SWE-bench Verified & Public benchmark-derived file-retrieval setting & 50 instances; 2,500 Qwen judgments & File retrieval only; no patch success \\
Synthetic v1/v3 & Controlled mechanism tests & 250 eval tasks each; 1,400/1,800 labels & Ranking sanity, removal, and budget effects \\
RepoBench-R & Real-code compression & 100 tasks; budgets 60--400 & Compression, not repair success \\
Judge/ranker pilots & Judge substitution diagnostics & 200 Opus-assisted labels; 710 Qwen-agreement candidates; 250 GPT-5.5 judgments & Surrogate and provider checks only \\
\bottomrule
\end{tabular}
\end{table}

\subsection{Methods Compared}

We compare CICL and CICL\_Distilled with NoContext, FullContext,
VanillaRAG, GraphMemory, SummaryMemory, SelfGeneratedExamples,
AutoContextKG, and OracleGoldContext. The last uses
gold ids only as an upper bound and is never employed as a selector
elsewhere. We additionally report seven CICL ablations removing individual
scoring components. In the synthetic result tables, CICL\_Distilled denotes
the learned student ranker family; rows marked CICL\_Dist.\ (linear) and
CICL\_Dist.\ (MLP) distinguish the linear and non-linear ranker variants.

\subsection{Metrics}

For each task we report simulated success rate, context
precision/recall/F1 against gold ids, mean reciprocal rank (MRR) of the
first gold id in the selection, average tokens consumed, and average tool
calls. For SWE-bench file retrieval, hit@1 measures whether the top-ranked file is
gold, coverage measures whether any gold file appears in the returned top 20,
R@k is the fraction of gold files retrieved in the top-$k$, averaged across tasks.
MRR@10 uses the first
gold file within the top 10, and gold rank reports the mean first-gold rank
over covered instances. In settings where paired task-level deltas are available, compression-suite
experiments additionally employ paired bootstrap
tests with $2{,}000$ resamples for delta metrics and report 95\% confidence
intervals; bootstrap pairs are matched per task id. The executable pilots
further report patch success and harmful-selection rates.

\subsection{Implementation Details}

The deterministic simulator, the heuristic CICL ranker, the AutoContextKG
selector, and the compression pipeline are implemented in pure Python with
fixed random seeds. The Opus-assisted annotation pipeline queries Claude Opus 4.7 with
provider-supported decoding settings using a counterfactual prompt template
that enforces the eight-field
JSON schema. The synthetic linear rankers are trained for $80$ epochs with
the pairwise logistic trainer (learning rate $0.08$, L2 $5\times10^{-4}$ in
the public script); the real-code pilot uses the same trainer for $100$
epochs. The MLP rankers use the same cleaned examples but optimise a BPR
pairwise loss for $120$ epochs with AdamW (learning rate $10^{-3}$, batch
size $256$, weight decay $10^{-4}$, dropout $0.1$). The Qwen3.5-9B QLoRA
judge is trained on two 16\,GB V100 GPUs and uses rank $r\!=\!8$,
$\alpha\!=\!16$, dropout $0.05$, target modules
\texttt{q\_proj, k\_proj, v\_proj, o\_proj}, effective batch size $16$
(batch $1 \times 16$ gradient accumulation steps), one epoch,
$\mathrm{lr}=2\times10^{-4}$, bf16 disabled (V100 compatibility), and 4-bit
NF4 base-weight quantisation. Local generation, training, and evaluation
scripts use fixed seeds. Decoding used deterministic or provider-default
settings where supported, and all external annotation runs keep archived JSON
outputs; because hosted LLM backends can drift, reported numbers are tied to
the released artifacts rather than to re-querying the provider.
The GPT-5.5 via Codex path uses the same code-retrieval prompt and eight-field
schema on the first five SWE-bench Verified instances; it is reported as a
small provider check, not as a full retrieval benchmark.

\section{Results}
\label{sec:results}

\subsection{SWE-bench Verified File Retrieval}
\label{sec:results-swe}

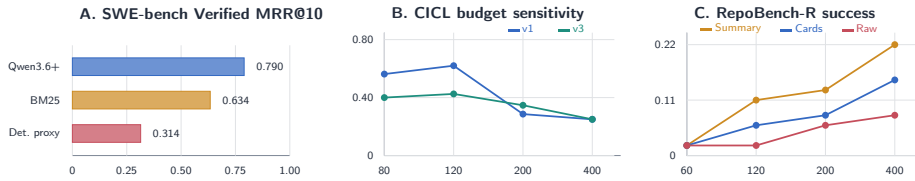
\begin{figure}[!tbp]
\centering
\begingroup
\definecolor{plotInk}{RGB}{34,43,58}
\definecolor{plotLine}{RGB}{82,94,113}
\definecolor{plotBlue}{RGB}{52,105,194}
\definecolor{plotTeal}{RGB}{30,142,126}
\definecolor{plotGold}{RGB}{205,139,34}
\definecolor{plotRose}{RGB}{197,77,91}
\begin{tikzpicture}[
  font=\sffamily\fontsize{7}{8.5}\selectfont,
  text=plotInk,
  axis/.style={draw=plotLine!75,line width=0.45pt},
  grid/.style={draw=plotLine!20,line width=0.35pt},
  lab/.style={align=center,inner sep=1pt},
  heading/.style={lab,font=\sffamily\bfseries\fontsize{8}{9.5}\selectfont},
  dotmark/.style={circle,inner sep=1.15pt},
  squaremark/.style={rectangle,inner sep=1.05pt},
  legend/.style={lab,anchor=west,inner sep=0pt}
]

\begin{scope}[xshift=0.10cm]
\node[heading] at (1.60,2.68) {A. SWE-bench Verified};
\node[lab] at (1.60,2.35) {File retrieval};
\draw[axis] (0,0) -- (3.20,0);
\foreach \x/\t in {0/0,0.8/0.25,1.6/0.50,2.4/0.75,3.2/1.00} {
  \draw[grid] (\x,0) -- (\x,1.97);
  \node[lab,anchor=north] at (\x,-0.07) {\t};
}
\foreach \name/\val/\y/\col in {
  Det. proxy/0.314/0.30/plotRose,
  BM25/0.634/0.90/plotGold,
  Qwen3.6-Plus/0.790/1.50/plotBlue} {
  \draw[fill=\col!70,draw=\col]
    (0,\y-0.10) rectangle ({3.20*\val},\y+0.10);
  \node[lab,anchor=west] at (0,\y+0.25) {\name};
  \node[lab,anchor=west] at ({3.20*\val+0.05},\y) {\val};
}
\node[lab] at (1.60,-0.58) {MRR@10};
\end{scope}

\begin{scope}[xshift=4.40cm]
\node[heading] at (1.50,2.68) {B. Budget sensitivity};
\node[lab] at (1.50,2.35) {CICL context F1};
\draw[axis] (0,0) -- (3.0,0);
\draw[axis] (0,0) -- (0,1.90);
\foreach \y/\t in {0/0,0.92/0.40,1.84/0.80} {
  \draw[grid] (0,\y) -- (3.0,\y);
  \node[lab,anchor=east] at (-0.06,\y) {\t};
}
\foreach \x/\t in {0/80,1/120,2/200,3/400} {
  \draw[grid] (\x,0) -- (\x,1.90);
  \node[lab,anchor=north] at (\x,-0.07) {\t};
}
\draw[plotBlue,line width=0.8pt] plot coordinates
  {(0,1.293) (1,1.426) (2,0.658) (3,0.575)};
\draw[plotTeal,dashed,line width=0.8pt] plot coordinates
  {(0,0.920) (1,0.978) (2,0.798) (3,0.575)};
\foreach \x/\y in {0/1.293,1/1.426,2/0.658,3/0.575}
  \node[dotmark,fill=plotBlue] at (\x,\y) {};
\foreach \x/\y in {0/0.920,1/0.978,2/0.798,3/0.575}
  \node[squaremark,draw=plotTeal,fill=white,line width=0.6pt] at (\x,\y) {};
\draw[plotBlue,line width=0.8pt] (0.72,2.06) -- (1.03,2.06);
\node[legend] at (1.09,2.06) {v1};
\draw[plotTeal,dashed,line width=0.8pt] (1.70,2.06) -- (2.01,2.06);
\node[legend] at (2.07,2.06) {v3};
\node[lab] at (1.50,-0.58) {Token budget};
\end{scope}

\begin{scope}[xshift=8.70cm]
\node[heading] at (1.50,2.68) {C. RepoBench-R};
\node[lab] at (1.50,2.35) {Simulated success};
\draw[axis] (0,0) -- (3.0,0);
\draw[axis] (0,0) -- (0,1.90);
\foreach \y/\t in {0/0,0.88/0.11,1.76/0.22} {
  \draw[grid] (0,\y) -- (3.0,\y);
  \node[lab,anchor=east] at (-0.06,\y) {\t};
}
\foreach \x/\t in {0/60,1/120,2/200,3/400} {
  \draw[grid] (\x,0) -- (\x,1.90);
  \node[lab,anchor=north] at (\x,-0.07) {\t};
}
\draw[plotGold,line width=0.8pt] plot coordinates
  {(0,0.16) (1,0.88) (2,1.04) (3,1.76)};
\draw[plotBlue,dashed,line width=0.8pt] plot coordinates
  {(0,0.16) (1,0.48) (2,0.64) (3,1.20)};
\draw[plotRose,densely dotted,line width=0.8pt] plot coordinates
  {(0,0.16) (1,0.16) (2,0.48) (3,0.64)};
\foreach \x/\y in {0/0.16,1/0.88,2/1.04,3/1.76}
  \node[dotmark,fill=plotGold] at (\x,\y) {};
\foreach \x/\y in {0/0.16,1/0.48,2/0.64,3/1.20}
  \node[dotmark,draw=plotBlue,fill=white,line width=0.6pt] at (\x,\y) {};
\foreach \x/\y in {0/0.16,1/0.16,2/0.48,3/0.64}
  \node[squaremark,fill=plotRose] at (\x,\y) {};
\draw[plotGold,line width=0.8pt] (-0.12,2.06) -- (0.14,2.06);
\node[legend] at (0.20,2.06) {Summary};
\draw[plotBlue,dashed,line width=0.8pt] (1.40,2.06) -- (1.66,2.06);
\node[legend] at (1.72,2.06) {Cards};
\draw[plotRose,densely dotted,line width=0.8pt] (2.46,2.06) -- (2.72,2.06);
\node[legend] at (2.78,2.06) {Raw};
\node[lab] at (1.50,-0.58) {Token budget};
\end{scope}

\end{tikzpicture}
\endgroup
\caption{Compact result overview. Panel A foregrounds the SWE-bench Verified
retrieval result; Panels B and C show the main controlled
budget trend and RepoBench-R compression boundary.}
\label{fig:results-overview}
\end{figure}

We begin with the open benchmark most likely to matter for coding agents:
SWE-bench Verified file-level retrieval. This setting does not measure patch
success, but it asks whether the selector can place the target file high enough
for a downstream agent to act. On 50 instances (Table~\ref{tab:swe}),
BM25/HybridRAG reaches hit@1 $0.58$ and MRR@10 $0.634$. The older deterministic
decision-utility proxy is weaker than BM25, but direct Qwen3.6-Plus judgments over the
top-50 file pool raise hit@1 to $0.78$ and MRR@10 to $0.790$, with 2500/2500
parseable judgments. The result is the paper's main positive real-code finding:
decision-aware judging provides a useful reranking signal when the judge can
read the candidate evidence well. The gain is in early ranking rather than
overall recall (R@10: $0.714$ versus BM25's $0.716$).
Figure~\ref{fig:results-overview} places
this open benchmark next to the controlled mechanism and compression trends.

\begin{table}[!htbp]
\centering
\caption{SWE-bench Verified file-level retrieval (50 instances), sorted by
non-oracle MRR@10. Direct Qwen3.6-Plus judgments re-rank top-50 candidates;
the metric is retrieval quality and not a patch result.}
\label{tab:swe}
\scriptsize
\setlength{\tabcolsep}{3.5pt}
\begin{tabular}{lcccccc}
\toprule
Method & Hit@1 & Coverage & R@5 & R@10 & MRR@10 & Gold rank \\
\midrule
Qwen CausalRerank & \textbf{0.78} & 0.80 & \textbf{0.712} & 0.714 & \textbf{0.790} & \textbf{1.03} \\
Qwen CausalHybrid & 0.70 & \textbf{0.82} & \textbf{0.712} & \textbf{0.733} & 0.733 & 1.88 \\
BM25              & 0.58 & 0.80 & 0.622 & 0.716 & 0.634 & 2.43 \\
HybridRAG         & 0.58 & 0.80 & 0.622 & 0.716 & 0.634 & 2.43 \\
Deterministic CausalRerank & 0.26 & 0.56 & 0.295 & 0.383 & 0.314 & 5.25 \\
HashedEmbedding   & 0.18 & 0.30 & 0.171 & 0.181 & 0.192 & 5.07 \\
Random            & 0.00 & 0.04 & 0.000 & 0.007 & 0.002 & 10.50 \\
OracleGoldContext & 1.00 & 1.00 & 0.975 & 0.984 & 1.000 & 1.00 \\
\bottomrule
\end{tabular}
\end{table}

We also ran the same prompt and schema through GPT-5.5 via Codex on the first five
SWE-bench Verified instances (250 candidate judgments). All judgments parsed;
on this five-instance slice, CausalRerank, CausalHybridRerank, BM25, and
HybridRAG all obtain hit@1 and MRR@10 of $0.60$. We use this only to check that
the schema can run through another provider; the 50-instance Qwen3.6-Plus run
is the main retrieval comparison.

\subsection{Ablation}
\label{sec:results-removal-budget}

Tables~\ref{tab:top-utility-removal} and~\ref{tab:budget-sweep} report the two
ablation diagnostics most directly tied to the selection mechanism: removing
selected evidence and varying the token budget. The clearest controlled signal
is the top-utility removal diagnostic. Removing the highest-scoring unit collapses
v3 context F1 from $0.245$ to $0.000$ and MRR from $0.980$ to $0.000$; random
removal is much less destructive (F1 $0.205$).

\begin{table}[!htbp]
\centering
\caption{Top-utility removal ablation on synthetic v3 (250 tasks, budget 400).
Removing the highest-utility unit collapses F1 and MRR to zero.}
\label{tab:top-utility-removal}
\small
\setlength{\tabcolsep}{6pt}
\begin{tabular}{lcccc}
\toprule
Condition & F1 & MRR & Precision & Recall \\
\midrule
CICL\_full              & 0.245 & 0.980 & 0.196 & 0.327 \\
CICL\_remove\_random    & 0.205 & 0.716 & 0.179 & 0.239 \\
CICL\_remove\_top       & 0.000 & 0.000 & 0.000 & 0.000 \\
\bottomrule
\end{tabular}
\end{table}

\noindent
The second controlled pattern is an inverted-U over token budgets
(Table~\ref{tab:budget-sweep}). CICL peaks at budget 120 on both suites
(v1 $0.620$, v3 $0.425$) and then decays as larger budgets admit more
distractors. On v3, VanillaRAG remains stronger, so the budget result is a
limited positive result rather than evidence of dominance.

\begin{table}[!htbp]
\centering
\caption{Context F1 across token budgets (250 tasks each). Rows are sorted by
non-oracle peak strength. CICL peaks at budget 120 on both suites; the
inverted-U is driven by precision loss at larger budgets. Bold marks each
row's peak.}
\label{tab:budget-sweep}
\scriptsize
\setlength{\tabcolsep}{4pt}
\begin{tabular}{lcccc|cccc}
\toprule
 & \multicolumn{4}{c|}{Synthetic v1} & \multicolumn{4}{c}{Synthetic v3} \\
Method & 80 & 120 & 200 & 400 & 80 & 120 & 200 & 400 \\
\midrule
AutoContextKG     & 0.706 & \textbf{0.850} & 0.671 & 0.481 & \textbf{0.481} & 0.406 & 0.441 & 0.305 \\
VanillaRAG        & 0.523 & 0.562 & \textbf{0.643} & 0.459 & 0.465 & \textbf{0.525} & 0.477 & 0.373 \\
CICL              & 0.562 & \textbf{0.620} & 0.286 & 0.250 & 0.400 & \textbf{0.425} & 0.347 & 0.250 \\
CICL\_Distilled   & \textbf{0.414} & 0.295 & 0.405 & 0.350 & \textbf{0.305} & 0.241 & 0.230 & 0.175 \\
\bottomrule
\end{tabular}
\end{table}

\subsection{Main Synthetic Ranking Results}
\label{sec:results-main}

Table~\ref{tab:main-synth} reports the standard budget-120 comparison, sorted
by average non-oracle F1 across the two suites. On v1, CICL is below
AutoContextKG but above VanillaRAG and GraphMemory. On v3, semantic distractors
change the ordering: VanillaRAG has the best F1 and AutoContextKG the best MRR,
while CICL remains competitive but trails VanillaRAG in F1.

\begin{table}[!ht]
\centering
\caption{Main 250-task evaluation at budget 120, sorted by average non-oracle
F1 across v1/v3. Bold = best non-oracle per metric.}
\label{tab:main-synth}
\small
\setlength{\tabcolsep}{4pt}
\renewcommand{\arraystretch}{0.92}
\begin{tabular}{lcccccc}
\toprule
 & \multicolumn{3}{c}{Synthetic v1 (structural)} & \multicolumn{3}{c}{Synthetic v3 (semantic)} \\
\cmidrule(lr){2-4} \cmidrule(lr){5-7}
Method & F1 & MRR & Tokens & F1 & MRR & Tokens \\
\midrule
AutoContextKG     & \textbf{0.850} & \textbf{0.984} & 94.3 & 0.406 & \textbf{1.000} & 98.4 \\
VanillaRAG        & 0.562 & 0.603 & 107.3 & \textbf{0.525} & 0.593 & 103.6 \\
CICL              & 0.620 & 0.817 & 117.8 & 0.425 & 0.896 & 117.2 \\
GraphMemory       & 0.511 & 0.654 & 116.8 & 0.303 & 0.617 & 117.0 \\
CICL\_Dist.\ (MLP)    & 0.328 & 0.277 & 111.2 & 0.334 & 0.273 & 113.0 \\
CICL\_Dist.\ (linear) & 0.295 & 0.302 & 110.6 & 0.241 & 0.226 & 113.7 \\
SelfGenExamples   & 0.138 & 0.338 & 98.6  & 0.124 & 0.366 & 101.9 \\
FullContext       & 0.020 & 0.037 & 115.0 & 0.027 & 0.045 & 114.0 \\
SummaryMemory     & 0.010 & 0.018 & 99.8  & 0.007 & 0.014 & 102.4 \\
NoContext         & 0.000 & 0.000 & 0.0   & 0.000 & 0.000 & 0.0   \\
\midrule
OracleGold        & 1.000 & 1.000 & 89.0  & 1.000 & 1.000 & 45.0 \\
\bottomrule
\end{tabular}
\end{table}

\subsection{RepoBench-R Compression Suite}
\label{sec:results-repobench}

Table~\ref{tab:repobench} reports the 100-task RepoBench-R Python compression
suite. At budget 120, memory-card compression improves over raw selection in
success ($0.02\!\to\!0.06$) and context recall ($+0.04$, 95\% CI
$[0.01,0.08]$, $p=0.016$ by matched paired bootstrap with $2{,}000$
resamples). Generic extractive summarisation is nevertheless stronger at the
same budget ($0.11$ success; summary-vs-card recall delta $+0.05$,
95\% CI $[0.01,0.10]$, $p=0.009$). In selected-then-compressed mode, memory
cards save $44.93$ tokens per query over raw selection (95\% CI
$[41.05,48.74]$, $p<10^{-3}$) with identical selected ids. The supported
result is therefore specific: cards can save tokens and improve raw budget
fitting, but generic summaries are the stronger baseline on this starter
slice.

\begin{table}[!ht]
\centering
\caption{RepoBench-R compression sweep over 100 tasks. Higher simulated
success and context F1 indicate better evidence selection. Compression ratio
is compressed-to-original token count; tokens and average units describe
budget use. Bold numbers mark the best rounded effectiveness result within
each budget.}
\label{tab:repobench}
\scriptsize
\setlength{\tabcolsep}{3.5pt}
\renewcommand{\arraystretch}{0.92}
\begin{tabular}{clccccc}
\toprule
& & \multicolumn{2}{c}{Effectiveness} &
\multicolumn{3}{c}{Efficiency and budget use} \\
\cmidrule(lr){3-4}\cmidrule(lr){5-7}
Budget & Method & Sim.\ success $\uparrow$ & Context F1 $\uparrow$ &
Tokens & Avg.\ units & Comp.\ ratio $\downarrow$ \\
\midrule
\multirow{3}{*}{60} & Summary & 0.02 & 0.009 & 55.2 & 2.63 & 0.26 \\
 & \textbf{Memory card} & 0.02 & \textbf{0.014} & 49.8 & 2.95 & 0.51 \\
 & Raw & 0.02 & 0.013 & 47.5 & 2.02 & 1.00 \\
\midrule
\multirow{3}{*}{120} & Summary & \textbf{0.11} & \textbf{0.030} & 115.3 & 5.31 & 0.27 \\
 & \textbf{Memory card} & 0.06 & 0.023 & 103.6 & 4.43 & 0.38 \\
 & Raw & 0.02 & 0.007 & 106.1 & 2.88 & 1.00 \\
\midrule
\multirow{3}{*}{200} & Summary & \textbf{0.13} & \textbf{0.022} & 193.2 & 8.90 & 0.24 \\
 & \textbf{Memory card} & 0.08 & \textbf{0.022} & 186.7 & 5.92 & 0.42 \\
 & Raw & 0.06 & 0.020 & 186.0 & 4.09 & 1.00 \\
\midrule
\multirow{3}{*}{400} & Summary & \textbf{0.22} & \textbf{0.030} & 335.1 & 14.51 & 0.19 \\
 & \textbf{Memory card} & 0.15 & 0.027 & 380.8 & 10.08 & 0.41 \\
 & Raw & 0.08 & 0.020 & 374.3 & 5.52 & 1.00 \\
\bottomrule
\end{tabular}
\end{table}

\subsection{Opus-Assisted Rankers and Qwen Selection Agreement}
\label{sec:results-qwen}

We collected 1\,400 Claude Opus 4.7-assisted counterfactual judgments on the v1
base tasks and trained lightweight rankers. The linear ranker reaches $0.939$
pairwise accuracy but remains below the heuristic on downstream F1; the MLP
improves over linear on both suites (v3 F1 $0.241\!\to\!0.334$). A separate
200-label RepoBench-R pilot gives only a small heldout signal, so we treat
real-code ranker training as preliminary. The Qwen3.5-9B QLoRA judge is used
only as an agreement diagnostic on 710 synthetic-distribution candidates
(Table~\ref{tab:qwen-agreement}), not as a standalone judge.

\begin{table}[!htbp]
\centering
\caption{Qwen3.5-9B QLoRA judge vs.\ Claude Opus 4.7 (710 candidates, 25 tasks).}
\label{tab:qwen-agreement}
\small
\setlength{\tabcolsep}{6pt}
\begin{tabular}{lc}
\toprule
Metric & Value \\
\midrule
JSON parse rate                       & 1.000 \\
Avg.\ top-5 Jaccard vs.\ Opus         & 0.592 \\
Avg.\ Spearman $\rho$ vs.\ Opus       & 0.379 \\
MAE (max across 5 numeric fields)    & 0.059 \\
\bottomrule
\end{tabular}
\end{table}

\subsection{Additional Diagnostics}
\label{sec:results-pilot}

A two-task toy patch-and-test pilot checks that selected context can pass
through an executable patch loop: CICL, VanillaRAG, and AutoContextKG reach
patch success $1.00$, while NoContext reaches $0.00$. Three Astropy
patch-generation smoke checks test formatting but are not official SWE-bench
passes. In harmful-context stress tests, CICL often selects stale units
($70.8\%$ on v1; $68.8\%$ on v3), but ranks the gold unit ahead of them
(harmful-before-gold $=0.00$). A structural audit of memory cards reports
field completeness $1.000$, action hints $1.000$, compression success $0.948$,
and token ratio $0.486$.

\section{Limitations}
\label{sec:limitations}

CICL has several scoped limitations. First, the score is
counterfactual-inspired rather than a formally identified causal effect; its
components are estimated by simulators, LLM judges, or rankers, and the
reported scorer uses fixed heuristic weights. Second, the main real-code result is SWE-bench Verified
file-level retrieval, not official patch success; the patch experiments are
small smoke checks. Third, judge quality matters: hosted models may drift, and
the Qwen-QLoRA result measures agreement on the synthetic Opus-assisted
distribution. Finally, typed cards can lose lexical detail, and strong
baselines remain competitive.

\section{Conclusion}
\label{sec:conclusion}

CICL treats context selection for tool-using LLM agents as a decision-aware
problem: useful context is evidence that should affect the next step, not
simply text that is similar to the task. The framework separates candidate
generation, decision-oriented scoring, and compact memory-card packaging under
an auditable judge schema.

Across the experiments, this framing yields a useful but bounded signal.
LLM-based reranking improves SWE-bench Verified file retrieval, removal
diagnostics show that top-scored units can be action-critical, and
selected-then-compressed cards reduce token use while preserving selected
evidence. The negative results are equally important: generic summaries can be
stronger for RepoBench-R context selection, compact rankers do not yet replace the heuristic
selector, and the evidence remains file-level rather than end-to-end repair
success. Future work should scale real-code annotations, learn adaptive utility
weights, combine structured cards with lexical snippets, and evaluate the
selector inside full agent rollouts.

\begin{credits}
\subsubsection{\discintname}
The authors have no competing interests to declare that are
relevant to the content of this article.
\end{credits}

\bibliographystyle{cicl_splncs04}
\bibliography{references}

\end{document}